%% file: paper.tex
\def\year{2018}\relax
\documentclass[letterpaper]{article} 
\usepackage{aaai18}  
\usepackage{times}  
\usepackage{helvet}  
\usepackage{courier}  
\usepackage{url}  
\usepackage{graphicx}  
\usepackage{fancyvrb}

\usepackage[utf8]{inputenc} 
\usepackage[T1]{fontenc}    
\usepackage{hyperref}       
\usepackage{url}            
\usepackage{booktabs}       
\usepackage{amsfonts}       
\usepackage{nicefrac}       
\usepackage{microtype}      

\title{Glass-Box Program Synthesis: A Machine Learning Approach}
\author{Konstantina Christakopoulou \\
University of Minnesota \thanks{Most of this work was done during an internship at MSR.}
\And
Adam Tauman Kalai \\ Microsoft Research}

\newcommand{\beq}{\begin{equation}}
\newcommand{\eeq}{\end{equation}}
\usepackage{enumitem}
\usepackage{float}
\usepackage[caption=false,font=normalsize,labelfont=sf,textfont=sf]{subfig}
\usepackage{caption}

\usepackage{graphicx} 
\usepackage{float}
\usepackage{amsmath}
\usepackage{enumitem}
\usepackage{amsfonts}

\usepackage{algorithm}
\usepackage[noend]{algpseudocode}
\usepackage{color}
\usepackage{float}

\newcommand{\ak}[1] {}

\usepackage{hyperref}

\nocopyright
\begin{document} 
\maketitle

\begin{abstract}
\input{new-abstract}
\end{abstract}

\section{Introduction}
\label{sec:intro}
\input{sec-intro}

\section{Related Work}
\label{sec:related}
\input{sec-related2}

\section{Proposed Framework}
\input{sec-overview}

\section{Experiments}
\label{sec:experiments}
\input{sec-experiments}


\section{Conclusions \& Future Directions}
\label{sec:concl}
\input{sec-concl}

\bibliographystyle{aaai}
\bibliography{paper.bib}
\end{document}

%% file: new-abstract.tex
Recently proposed models which learn to write computer programs from data use either input/output examples or rich execution traces. Instead, we argue that a novel alternative is to use a \emph{glass-box} loss function, given as a program itself that can be directly inspected. Glass-box optimization covers a wide range of problems, from computing the greatest common divisor of two integers, to learning-to-learn problems. 

In this paper, we present an intelligent search system which learns, given the partial program and the glass-box problem, the probabilities over the space of programs. We empirically demonstrate that our informed search procedure leads to significant improvements compared to brute-force program search, both in terms of accuracy and time. For our experiments we use rich context free grammars inspired by number theory, text processing, and algebra. Our results show that (i) performing 4 rounds of our framework typically solves about 70\% of the target problems, (ii) our framework can improve itself even in domain agnostic scenarios, and (iii) it can solve problems that would be otherwise too slow to solve with brute-force search. 


%% file: sec-intro.tex
For computers to program computers, we must first address how programming problems will be represented and how performance will be evaluated. In the field of program synthesis, the two main approaches for specifying problems are: (a) by examples \cite{gulwani2012spreadsheet}, in which a number of example input-output pairs $(x_i,y_i)$ are provided as input and the goal is to output a function $f$ satisfying $f(x_i)=y_i$ while possibly minimizing other criteria (e.g., being short); and (b) by specification \cite{manna1980deductive}, in which a formal specification in some particular language is given.  More generally, there is a utility (usefulness score) for any synthesized program. Assuming this utility function can be written as a program-scoring program, we propose the approach of giving the synthesis direct access to the scoring-program's source code: (c) program synthesis as optimizing the {\em glass-box}\footnote{Ironically, the term {\em white box} is commonly used to indicate transparency even though white boxes are not necessarily transparent. Hence, we use {\em glass box}.} program scoring objective that will be used to evaluate it. In this paper, we illustrate the potential of the glass-box program synthesis approach by designing a system that learns to synthesize programs that maximize the corresponding utilities of  various glass-box objectives. 
%

\begin{figure}
\begin{center}

\begin{tabular}{|c|c|}
 \hline
  {\bf Input $x$} & {\bf Output $y$}\\\hline
  BUBQJ & B\\\hline
  ZMFXI & Z\\\hline
  NEJOL & N\\\hline
  RVOII & I\\\hline
  \end{tabular}
\begin{Verbatim}[frame=single, fontsize=\footnotesize]
def score(x: string, y: char):
    return x.count(y)
sum([score(x, f(x)) for x in tests()])
\end{Verbatim}
\end{center}
\caption{Two representations of the problem of finding the most frequent character in a string. \textbf{Top}: example input-output $(x,y)$ pairs. \textbf{Bottom}: 
glass-box representation, summing scores on test strings. The score is the number of occurrences of $y$ in $x$, and \texttt{tests()} randomly generates strings.
\label{fig:example}}
\vspace{-0.2cm}
\end{figure}
\subsubsection{Glass-box program synthesis.}
To better understand the glass-box representation, consider a program synthesis contest. In programming contests among humans, problems are often described in English and scored automatically by automated {\em scoring programs} based on their output on certain inputs (and other factors such as runtime and time of submission). Certainly, it is difficult for computers to understand English descriptions, so instead, we propose describing the problem to the synthesis system through the source code of the scoring program. In this ``glass-box'' model, there is no need for a separate problem description or examples -- just the scoring program. As illustrated in Figure \ref{fig:example}, if the problem was to find the most frequent character in a string, the score of a program outputting a character $y$ on a string $x$ would be the number of occurrences of $y$ in $x$ 
and the total score would be the sum of its scores over some randomly generated strings. Note that to specify the problem by input-output examples (Figure \ref{fig:example} top), one needs to solve the problem on several examples, and there may be ambiguities in that there may be multiple different functions $f$ mapping $x$ to $y$. 

In casting the problem of programming as optimizing a glass-box program-scoring program, one must write a program that precisely defines the score/utility of the synthesized program. However, this is arguably a necessary step to posing a problem in general, not only for programming contests. The term {\em glass-box} contrasts with black-box access. Black-box access would mean the ability to score arbitrary programs without any other access to the scoring program. Glass-box access is of course at least as powerful as black-box access because one can run the scoring program itself. 
A few cases where glass-box program synthesis can be applied are:
\begin{itemize}[noitemsep]
\item Traditional optimization problems such as linear programming or the Traveling Salesman Problem (where the objective is the length of the tour). 
\item Number theory problems such as Greatest Common Divisor (GCD) or factoring. These problems can be efficiently scored because it is easy to verify factors and  primality. 
\item Programming by example (PBE). In this case, a program $P$ is scored by its accuracy mapping fixed inputs $x_i$ to outputs $P(x_i)=y_i$, combined with a regularization term, e.g., $-\lambda(\text{program length})$, to prevent over-fitting. 
\item Optimizing an algorithm's performance in simulation. An example would be designing a network protocol to be evaluated in a network simulator.  In this case, the scoring program could measure performance in a large network.
\item Meta-optimization, learning to learn to learn. 
The problem of synthesizing program synthesis can itself be posed in the form of a meta-scorer that generates problems (i.e., scorers) from various domains, runs the candidate synthesizer on these problems, and averages the resulting program scores.  
\end{itemize}


\subsubsection{Learning to synthesize solutions by synthesizing problems.}
After defining the representation and evaluation of a programming problem, we demonstrate the feasibility of this approach through a system that learns to synthesize programs. Just as athletes do various exercises to improve performance at a sport, a program synthesis system may improve by practicing and learning from synthesizing solutions to various problems. Menon et al.\ (\citeyear{menon2013machine}) introduced a Machine Learning (ML) approach to PBE synthesis that learns to synthesize across problems. Because their repository of real-world problems was relatively small, they selected a small number of features suited to text-processing PBE. 

To get around this shortage of data, we generate our own problems which we then use to practice synthesizing solutions. In particular, once we have a set of practice problems, we iteratively find improved solutions to these problems by interleaving search and training a logistic regression-based model which guides the search intelligently, following  Menon et al.\ (\citeyear{menon2013machine}). Recall that in glass-box synthesis, a problem is a scoring function, so we immediately know how to score any synthesized program. 
A similar approach of creating artificial problems was introduced independently by Balog et al.~(\citeyear{balog2016deepcoder}). Two key differences between this work and ours is that they perform PBE synthesis while we perform glass-box synthesis, and that they utilize deep learning, while we make use of multi-class logistic regression. 

\subsubsection{Experiments.}We show that in practice, it is possible to synthesize solutions for a number of problems of interest, such as the GCD, that would be prohibitively slow with a naive approach. Also, we show that our Glass-Box Program Synthesis system  (GlassPS), can improve itself even in a domain agnostic framework, where a union of grammars from various domains is considered; although better results can be achieved with domain-specific grammars.

\subsubsection{Contributions.} 
In this paper, we formalize learning to synthesize successful solutions to programming problems as a machine learning problem, using glass-box optimization. Our main contributions are three-fold: 
\begin{enumerate}[noitemsep]
\item We introduce glass-box program-scoring programs as a novel alternative for specifying problems.
\item We formalize a machine learning framework for glass-box optimization by synthesizing practice problems and learning patterns among the  problems and the as of now discovered solutions. 
\item We present experiments that demonstrate the ability of our framework to learn to generate well-formed python programs across domains.
\end{enumerate}

The rest of this paper is organized as follows. After discussing related  work, we introduce key concepts needed to understand our approach, and then present the details of our proposed learning to write programs framework GlassPS. 
Then, we present experimental results that measure the empirical performance of GlassPS in a range of problem domains.  

%% file: sec-related2.tex
 

Learning to write computer programs has recently received a lot of attention from multiple viewpoints. Here we mention only a few notable related works due to space constraints. 

In \emph{Programming by Example} (PBE), a program synthesis system attempts to infer a program from input/output (I/O) example pairs, searching for a composition of some base functions. PBE has had success in various domains \cite{gulwani2012synthesis}, one notable example being "Flash Fill" for string manipulation in Microsoft Excel \cite{gulwani2012spreadsheet}. In (Raza et al.~\citeyear{raza2015compositional}) the end-user can give both I/O pairs and a natural language description of the task.

Recent advances in deep learning and the augmentation of deep networks with end-to-end trainable abstractions \cite{graves2016hybrid},  such as Neural Turing Machine (Graves et al.~\citeyear{graves2014neural}), Hierarchical Attentive Memory \cite{andrychowicz2016learning} and Neural Stack \cite{joulin2015inferring}, have given rise to the \emph{neural programming} paradigm (Neelakantan et al.~\citeyear{neelakantan2015neural}; Zaremba and Sutskever~\citeyear{zaremba2014learning}). Most of these works are trained using I/O examples, except for (Reed and de Freitas~\citeyear{reed2015neural}; Cai et al.~\citeyear{cai2017making}) where the model (neural programmer interpreter) is trained with rich supervision execution traces. 



The works of \cite{menon2013machine,yessenov2013colorful,balog2016deepcoder,dechter2013bootstrap,devlin2017robustfill,parisotto2016neuro}  combine the learning-to-program approaches of machine learning and program synthesis to perform \emph{guided search} over the space of synthesized programs.




Other successful views have been the \emph{probabilistic programming} perspective, i.e., representing a program as a generative probabilistic model (Lake at al.~\citeyear{lake2015human}), and the \emph{programming via specification} approach \cite{solar,gaunt2016terpret}, i.e., specifying a partial program `sketch' capturing the high-level structure of the implementation  and letting the computer synthesize the low-level details. 

 
\subsubsection{Relationship to Other Works.} Our work, using the perspective of machine learned program synthesis, introduces the novel glass-box introspection along with contextual features to inform the search. The two closest works to ours are \cite{parisotto2016neuro} and \cite{balog2016deepcoder}. The key differences to our work are: (i) they use I/O examples to condition the search, while we propose and use the glass-box problem representation, (ii) they use deep networks, while we use logistic regression. 
Also, while in our work and \cite{parisotto2016neuro}, problem-specific learned weights and a partial program representation guide the search, in \cite{balog2016deepcoder} a separate model has to be learned per task. 

Another factor differentiating the various works  is the expressiveness of the Domain Specific Language used. 
While many learning-to-program works demonstrate program-writing for single domains like string processing, our approach can generate code in a general-purpose programming language covering various domains such as number theory, strings, root finding; which can open up interesting possibilities for general problem solving (Mikolov et al.~\citeyear{mikolov2015roadmap}). 

Similar to \cite{balog2016deepcoder}, in our approach we utilize the thus far found problem-solution pairs to inform the continuous learning of our system; hence, our work can be put in the context of lifelong learning \cite{gaunt2016lifelong}.

Having said this, our goal in this paper is \emph{not} to compete with existing PBE systems or Neural Programmer Interpreter ones. Instead, we wish to provide glass-box representation as an alternative \emph{tool} to guide the program generation.

%% file: sec-overview.tex

\subsection{Key Concepts}
To formalize the problem of learning to synthesize programs as solutions to glass-box optimization problems, we start with a discussion of high-level concepts: 


\noindent \textbf{Program.} A program $P$ computes a function $p: \mathcal{X} \rightarrow \mathcal{Y}$ where $\mathcal{X}$ is a set of inputs and $\mathcal{Y}$ is a set of outputs. 
We distinguish the program $P$ from the function it computes $p$ because two different programs may compute the same function. In GlassPS, program input $\mathcal{X}=\mathcal{O}$ is the set of python objects (including numbers, strings, arrays, and functions) and program output $\mathcal{Y}=\mathcal{O}\cup\{\bot\}$ is the set of objects plus the special symbol $\bot$ that indicates that the program crashed or did not produce an output in the allotted time. \\
\textbf{Glass-box problems.}
Glass-box synthesis is defined over a set of {\em problems}. Each problem is represented by a glass-box scoring program $P \in \mathcal{P}$. A problem $P$ computes a function $p:\mathcal{S} \rightarrow \mathbb{R}$, which measures the score of the solution $S$ to the problem $P$, i.e., $p(S)$. \\
\textbf{Synthesizer.}
A {\em synthesizer} $Z:\mathcal{P} \rightarrow \mathcal{S}$ generates a solution program  based on the program with which it will be scored. Hence, the goal of a synthesizer is to attempt to maximize score:
$Z(P) \approx \arg\max_{S \in \mathcal{S}} p(S).$

Importantly, $Z$ takes the scoring-program $P$'s source code as glass-box input. Though this can be used to simulate black-box access to $p$ by generating I/O examples, the synthesizer $Z$ can potentially achieve higher scores than with black-box access alone. \\
\textbf{Solutions.} We use $\mathcal{S}$ to denote the set of programs output by the synthesizer, which we refer to as {\em solutions}. Hence, note that both problems and solutions are programs. \\
\textbf{Grammars.}
A program can be expressed as a composition of building blocks; these blocks constitute the rules of a context-free \emph{grammar} (CFG) that allows for recursive and compositional structure. We denote the set of grammar rules for solution programs by $\mathbb{S}$ and for problem programs by $\mathbb{P}$. 

The solution CFG $\mathbb{S}$ GlassPS uses, generates program trees that are converted to strings and are then evaluated by the python interpreter. It includes rules such as,
\begin{align*}
\text{R1:~} E \rightarrow (E + E), ~~& \text{R2:~}E \rightarrow (\text{lambda $x$: }E) \\
 \text{R3:~} E \rightarrow (E)\text{.lower()}, ~~ &\text{R4:~} E \rightarrow 1, ~~ \text{R5:~}E \rightarrow x
\end{align*}
For instance, (R1) generates code to add numbers, concatenate strings, or combine any two objects supported by the python $+$ operator. Rule (R2) creates a function of one variable, $x$. Rule (R3) converts a string to lower-case. The grammar supports iteration through recursion. We avoid halting issues \cite{skiena} by bounding the total number of routine calls allowed. For simplicity, $\mathbb{S}$ does not have types and uses only one non-terminal; leaving it to the learning system to \emph{learn} to generate programs that do not raise exceptions. Note that our CFG formulation does not support us directly ``reaching in'' to the scoring program, e.g., to extract constants, though such functionality could be added to our system.

The problem CFG $\mathbb{P}$ GlassPS uses, also generates code to be evaluated by the python interpreter. $\mathbb{P}$ contains multiple non-terminals, roughly grouped by python type, so as to facilitate generating well-formed scoring programs.

\noindent \textbf{Program Tree.} Glass-box/Solution programs are derived from the corresponding CFG ($\mathbb{P}/\mathbb{S}$) and are represented as rooted trees in which each node is associated with a rule from the CFG. 
Problem/Solution trees are constructed top-down probabilistically, sampling from the rule probabilities of $\mathbb{P}$ or $\mathbb{S}$ respectively. 
The choice of trees for representing programs is convenient, as it is easy to extract features from trees for machine learning purposes. 
 



\subsection{Learning from (practice problems, solutions).} 
It is natural to try to {\em learn} to synthesize programs based on a collection of problem-solution pairs. To do so, one would ideally have access to a large repository of samples of problem and solution programs. Menon et al.~(\citeyear{menon2013machine}) provide a small set of problem/solution pairs for text processing PBE. Since the set is relatively small, they use domain knowledge to construct a small number of hand-coded features for learning. 

\begin{figure}[h]	
	\centering
\includegraphics[scale=0.5]{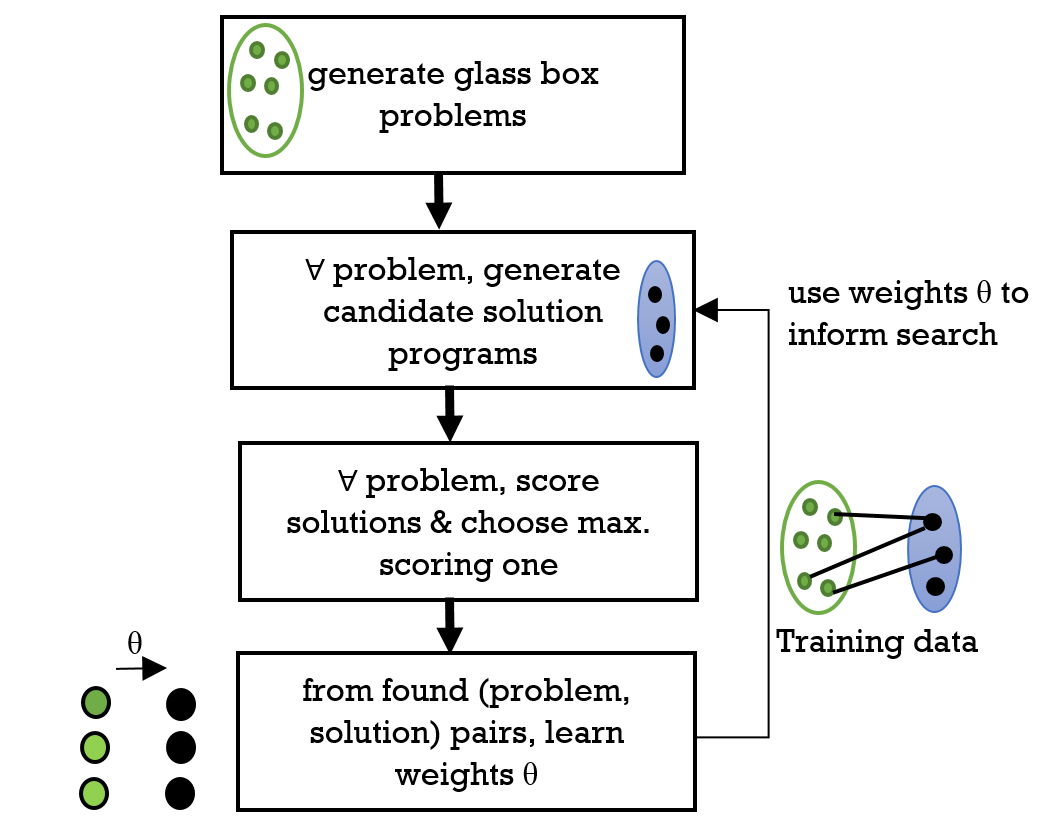}%
\caption{Framework Overview. }
\label{fig:overview}
\end{figure}

Instead, similarly to  Balog et al.~(\citeyear{balog2016deepcoder}), we synthesize practice problems of our own, and synthesize solutions to these problems. In our glass-box synthesis approach, this amounts to synthesizing scorers, i.e., the glass-box problems. For each such scorer, we  synthesize a number of solution programs specific to that scorer and choose the highest scoring program. We then learn from this collection of problem-solution pairs to improve the model used in synthesis. We iteratively find improved solutions to these problems by interleaving search and training a model that helps guide the search intelligently. 
An overview of our framework is shown in Figure \ref{fig:overview} and the specifics are described next. 

\textbf{Synthesizing practice problems.} Since glass-box problems are programs themselves, they can be represented as program trees and they are synthesized by randomly expanding nodes from the CFG $\mathbb{P}$ (uniform probabilities), on which the set of problems can be expressed. 
Duplicate problems are removed, and the synthesized problems are divided into training and test sets, with a 90-10 random train/test split. The training (practice) problems are denoted $P_1, \ldots, P_m \in \mathcal{P}$ and the test problems $T_1, \ldots, T_n$. 


\textbf{Challenge Problems.} Apart from the practice problems that we synthesize, and the set of test problems held out from this pool of synthesized problems for validation, we also use ``challenge'' problems. 
Challenge problems are ten problems that we created manually, written in terms of  $\mathbb{P}$, and intended to be ``representative'' of programming challenges that are naturally expressed as glass-box synthesis problems. 
These problems are given in the Experiments Section.  

\textbf{Synthesizing solutions.} For each practice/test/challenge problem, a fixed number of candidate solution program trees are synthesized top-down probabilistically using the CFG $\mathbb{S}$, and the best one, according to the scoring function of the problem at hand, is chosen. 

Every rule of the CFG $\mathbb{S}$ is associated with a probability. A program,  which is a collection of rules, has as probability the product of probabilities of the rules that comprise it. Before learning, the probabilities of the rules in $\mathbb{S}$ are equal (uniformly random expansion). The purpose of using machine learning is to learn the problem-specific rule probabilities of $\mathbb{S}$ successfully, so as to make the successful solution programs more probable, and thus easier to find. In what follows, we specify how to formalize the \emph{learning to write programs that optimize glass-box functions} as a machine learning problem.  


\textbf{Input Features.} As input to the machine learning classifier, we use two types of features: {\em glass-box problem features} and {\em context features}. The \emph{problem} features used for learning are a ``bag-of-rules'' representation of the problem. That is to say, for each  problem rule in $\mathbb{P}$, there is a feature for its number of occurrences in the problem's program tree. By {\em context} we refer to the partial candidate program created so far. Specifically, since synthesis is top-down, we mean the path of rules from the root of the generated tree to the current node whose rule is to be assigned.  For simplicity, to represent the context, we only use the one-hot-encoding of the parent node's rule and the one-hot encoding of what the current node's child index is (i.e., is it the first/second/etc. child of the parent node). 
Thus, the one-hot-encoding vectors of the glass-box problem, the parent-node-rule and the child-index are concatenated, and comprise the input $\phi$ to the learner.\footnote{Although the bag-of-word representation may seem simplistic, as it loses the program structure information, preliminary experiments using richer representations, such as sequence models, did not seem to offer much value to trade-off the computational expenses.} 

\textbf{Learning rule probabilities of $\mathbb{S}$.} For every node of a candidate solution program tree, we want to predict using the derived features $\phi$ described above, which rule of $\mathbb{S}$ is the most probable. Thus, the problem of program inference, i.e., searching over the space of programs expressed by CFG $\mathbb{S}$, can be reduced to a multi-label classification problem, as a solution program $S$ is a collection of rules that are \emph{simultaneously} present. This can be further reduced to multi-class classification, predicting for each class-rule separately whether it should be present or not. The number of classes equals $|\mathbb{S}|$.

We represent the target label of the correct rule in the successful solution program $S_{P}^*$ for glass box problem $P$ as a one-hot-encoding vector $y \in \mathbf{R}^{|\mathbb{S}|}$ with 1 in the the next node rule present in $S_{P}^*$, and 0 in the other $|\mathbb{S}|-1$ entries. 

\begin{figure}
\includegraphics[scale=0.25]{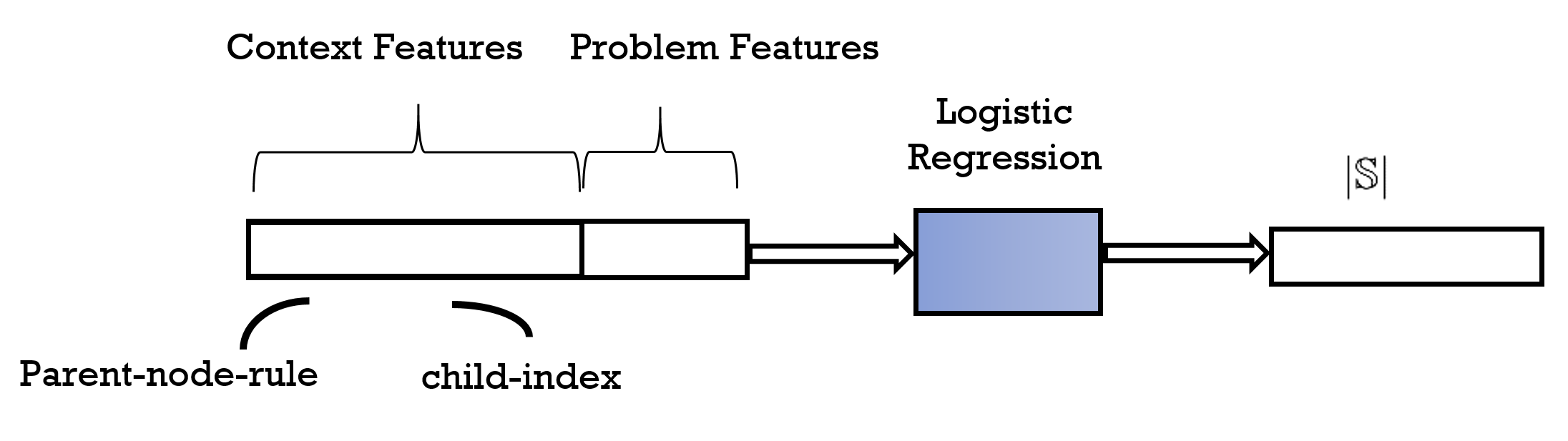}%
\caption{Mapping input features $\phi$ to class/rule $y \in \mathbf{R}^{|\mathbb{S}|}$.}
\label{fig:LR}
\end{figure}

Given the general formulation, any multi-class classifier is applicable. In this work, for the purposes of a proof-of-concept illustration, we use a multi-class logistic regression model with parameters $\Theta$; in particular, we learn $|\mathbb{S}|$ such parameter vectors, one for each class. An illustration of the mapping of features to solution rule is shown in Figure \ref{fig:LR}.

It is important to note that the parameters $\Theta$ are learned based on a constantly updated dataset consisting of the thus far found successful (problem, solution) program pairs. Given the featurization procedure described, each found successful solution program with e.g. $r$ nodes contributes $r$ samples for the learning of the classifier; for each of these samples the input features will differ as the context features will change.

\textbf{Scoring Solution Programs.}
In order to include a successful problem-solution pair in the constantly updated training dataset, a notion of success needs to be specified. Recall that a solution program is successful if it maximizes the score of the corresponding problem. Although our framework is entirely capable of handling arbitrary continuous scoring functions, the nature of the problems induced by the grammars used in our experiments is such that all scoring programs return scores in $\{0,1\}$ except for a solution that throws an exception, in which case the score is $-1$. This makes it easier to evaluate our system as the average score indicates the fraction of problems optimally solved. With iterations of learning, given that $\Theta$ are used to inform the problem specific search over $\mathbb{S}$, the poor-scoring programs become less likely. 

\textbf{Learning to write typed programs.} While it would be nice if GlassPS generated a subset of python programs, in fact it initially generates many nonsensical (un-typable) programs because it uses only a single terminal $E$. Thus, unsurprisingly, at first the grammar mostly generates programs that raise exceptions by doing things like trying to lower-case the number 1. As a result, our learning system, utilizing the of so far found pairs of glass box problems-solutions, will progressively learn to compose well-formed programs, in addition to learning to solve problems. 

\subsection{Algorithmic Procedure}
To summarize our framework, a formal description of the procedure followed is given in Algorithm \ref{alg:learn}. Our system operates on iterations $j=1, \ldots, T$. At $j$ round, GlassPS calls the SOLVE module, in order to attempt to solve the $T_1, \ldots, T_n$ test problems, which are synthesized based on $\mathbb{P}$. In SOLVE, the system generates its own practice problems $P_1, \ldots, P_{m}$ from the CFG $\mathbb{P}$, using uniform probabilities. To solve these generated practice problems, the system calls the TRAIN module. The goal of TRAIN is to a) construct a training dataset of problems - solutions, so that b) the learner's parameters $\Theta$ are learned. To achieve a), the system uses the parameters of the previous round's logistic regression model, to guide how the program trees of the solution programs will be built (SEARCH). Specifically, for every node of the under construction solution program tree, features are extracted via the module FEATURIZE (discussed in the previous Section), which creates the bag-of-word features $\phi$ for the problem and context. Given these features, the current learner's model is used to predict, what the probability of every rule of the solution CFG $\mathbb{S}$ is (inside the module LR-predict). These inferred probabilities are used as per rule weights to perform weighted sampling for which rule should be next in the candidate program tree. This is how the learning guides the search over programs. 

Via this procedure, for every  problem, a set of candidate solution programs is constructed. Each candidate program $S_i$ is scored by the corresponding scoring program $P(S_i)$. The programs which successfully solve the respective glass-box problems, i.e., with the maximum score using shortest length to break the ties, are used to construct a (problem, context) $\rightarrow$ solution rule training data set, calling again the FEATURIZE routine for the $\phi$ representation of (problem, context). This constructed training dataset is then used to learn a new logistic regression model, which will be used to find solutions for the test and challenge problems, and which will be subsequently used in the next $j+1$ learning round to guide the search over the candidate solution programs. 

\begin{algorithm}[h]
\caption{Learning algorithm for glass-box synthesis}\label{alg:learn}
\begin{algorithmic}[0]
\Procedure{Solve}{$m, T_1, \ldots, T_n$}\Comment{learn \& solve}
\For{$i=1$ to m}:
\State $P_i \gets \Call{GenRandomPractice}{}$
\EndFor
\State $\theta \gets \Call{Train}{P_1, P_2, \ldots, P_m}$
\For{$i=1$ to n}:
\State $S_i \gets \Call{Search}{T_i, \theta}$
\EndFor
\State \textbf{return} $S_1,S_2, \ldots, S_N$
\EndProcedure

\end{algorithmic}

\begin{algorithmic}[0]
\Procedure{Train}{$P_1, \ldots, P_m$}\Comment{fit $\theta$}
\State $\theta \gets \theta_0$
\For{$j=1$ to num-epochs}
\For{$i=1$ to $m$}
\State $S_i \gets \Call{Search}{P_i, \theta}$
\EndFor
\State Featurize and train LR on all nodes in $\langle(P_i, S_i)\rangle_{i=1}^m$
\State Update $\theta$
\EndFor
\State \textbf{return} $\theta$
\EndProcedure
\end{algorithmic}

\begin{algorithmic}[0]
\Procedure{Search}{$P$, $\theta$}\Comment{solve a problem}
\For{$i=1$ to num-candidates}
\State $S_i \gets \Call{Node}{P, \text{"root"}, \theta}$
\EndFor
\State \textbf{return} $S_i$ with greatest score $P(S_i)$
\EndProcedure
\end{algorithmic}

\begin{algorithmic}[0]
\Procedure{Node}{$P$, $c$, $\theta$}\Comment{node from prblm, context}
\State $\phi \gets \Call{Featurize}{P, c}$
\State $i \gets \text{weighted-sample}(\text{LR-predict$_\theta$}(\phi))$ \Comment{rule $r_i$}
\State $\text{children} = []$
\For{$j=1$ to number-of-rule-children($r_i$)}
\State $\text{children.append}\left(\Call{Node}{P, (r_i, j), \theta}\right)$
\EndFor
\State \textbf{return} new node with rule $r_i$ and children
\EndProcedure
\end{algorithmic}

\end{algorithm}

\subsection{Illustrative Example: how guided search works}
\label{subsec:ill}
Before we move to our experimental results, to better understand how our learning framework works, let us demonstrate how we could successfully synthesize the solution program of computing the greatest common divisor (GCD) of two positive integers $m, n$ using our framework GlassPS. 

The particular glass-box scoring function for GCD is 
$
\sigma(m, n, y) = \begin{cases} 0 & \text{if }m \text{ mod } y\neq 0\\
0 & n \text{ mod } y\neq 0\\
y & \text{otherwise}
\end{cases}$
, which has maximum score $y$, achieved when $y$ is a factor of both $m$ and $n$. 
In python, this program can be written succinctly as \texttt{lambda m, n, y: ntprog(-mod(m, y) or (-mod(n, y) or y))}, where \texttt{ntprog} computes the score over a domain of $m,n$ pairs. It also implements an early stopping optimality criteria, which improves efficiency without changing the behavior of the algorithm. 

The successful solution program that we are after $S^*$ is \texttt{def f(m, n): f(mod(n, m), m) if n else m))}, i.e., a recursive form of Euclid's GCD algorithm. 
\begin{table}
\begin{minipage}{0.3\textwidth}
{\small
 \begin{tabular}{|ll|} 
 \hline
 Toy Glass-box Problem CFG $\mathbb{P}$&\\ 
 \hline
 W1: Loss $\rightarrow $ ntprog($Obj$)   & W6: $NT \rightarrow y $ \\
 W2: $Obj \rightarrow $ lambda $m, n, y: NT$ & W7: $NT \rightarrow n $\\ 
 W3: $NT \rightarrow $  $(NT$ or $NT)$ & W8:  $NT \rightarrow 0 $  \\
 W4: $NT \rightarrow $  mod($NT, NT$) & W9: $NT \rightarrow -NT $  \\
 W5: $NT \rightarrow m $ & W10: $NT \rightarrow NT < NT $  \\
 \hline 
 Toy Solution Program CFG $\mathbb{S}$& \\ 
 \hline
  R1: $E \rightarrow $rec(lambda $m, n: E$) & R9: $E \rightarrow 0 $ \\
  R2: $E \rightarrow $callrec($E,E$) & R10: $E \rightarrow 1 $ \\ 
  R3: $E \rightarrow E$ if~ $E$ else $E$ & R11: $E \rightarrow \log(E)$  \\
  R4: $E \rightarrow $mod($E, E$) & R12: $E \rightarrow \text{arctan}(E) $ \\
  R5: $E \rightarrow m $ & R13: $E \rightarrow E.upper()$ \\
  R6: $E \rightarrow n $ & R14: $E \rightarrow $ [$E$ for $i$ in $E$] \\ 
  R7: $E \rightarrow E + E $ & R15: $E \rightarrow set(E)$ \\
  R8: $E \rightarrow abs(E) $ & R16: $E \rightarrow (E, E)$  \\
\hline
\end{tabular}
}
\end{minipage}
\caption{Toy grammars for illustration purposes. \texttt{callrec} refers to the call of a function recursively, \texttt{rec} is the outer function to be called recursively, \texttt{ntprog} is a function for early-stopping based on evaluations of the synthesized program in various argument values. $\mathbb{P}$ contains many terminals ($Loss, Obj, NT$), while $\mathbb{S}$ contains a single terminal $E$.}
\label{tab:ill}
\end{table}

For illustration purposes, let us consider the toy CFGs shown in Table \ref{tab:ill}, which are actually subsets of the grammars used in our experiments. They contain rules typically useful for number theory problems. 

Calling the FEATURIZE module of Algorithm 1, the glass-box problem of GCD is written as a one-hot-encoding vector $\mathbf{x}_{\text{prob}}^{\text{GCD}} \in \mathbf{R}^{|\mathbb{P}|}$, with non-zero values $W1: 1, W2: 1, W3: 2, W4: 2, W5: 1, W6: 3, W7: 1, W9: 2$, where the values are the number of occurrences of the respective rule.
 
Recall that to decide which rule will become the next program tree node, a learned Logistic Regression (LR) model based on the so far found pairs of problems-solutions will be used in \emph{inference} mode, to predict given the problem \& context features $\phi$, which target class $y$ (i.e., rule of the solution CFG) is the most likely. Thus, to  construct the successful finding-GCD program tree $S^*$ top-down, guided search will proceed by \emph{sampling rules} from $\mathbb{S}$ for the next node  using the predicted LR probabilities:

\begin{enumerate}[noitemsep]
\item For the root node: the features $\phi$ will be $\mathbf{x}_{\text{prob}}^{\text{GCD}}$ concatenated with an all-zero vector, as the root does not have a parent. The target \emph{sampled} class $y$, i.e., the rule with the maximum sampled predicted probability from LR, should be $R1$ (one-hot-encoding with 1 in the entry corresponding to $R1$). 
\item For the second node, $\phi$ will be $[\mathbf{x}_{\text{prob}}^{\text{GCD}}, 
$ $\mathbf{x}_{\text{parent}}$, $\mathbf{x}_{\text{child}}]$, where $\mathbf{x}_{\text{prob}}^{\text{GCD}}$ is the same as for the previous node, $\mathbf{x}_{\text{parent}} \in \mathbf{R}^{|\mathbb{S}|}$ is the one-hot encoding vector with 1 in $R1$ (as $R1$ is the rule of the parent node), and $\mathbf{x}_{\text{child}}$ is the one-hot vector with 1 in the first entry, as this is the first child. The sampled target class from the predicted probability vector should be $R3$ (the if rule). 
\item Continue for each program tree node, until no more nodes are to be expanded, i.e., they are leaf nodes, and the entire $S^*$ program tree is constructed.
\end{enumerate}
In short, the LR model should learn that the rules $R1, R2, R3, R4, R5, R6$ should have high probability under the $\mathbf{x}_{\text{prob}}^{\text{GCD}}$ and the certain parent and child index contexts. When finding $S^*$ corresponding to the glass-box GCD $P_{GCD}$, the $(P_{GCD}, S^*)$ pair is added to the training dataset used to learn next round's LR.

%% file: sec-experiments.tex

\subsubsection{Setup.} We use as evaluation metric the fraction of test problems successfully solved. The sets of practice/train and test problems do not change throughout the experiment. Programs were limited to be at most size 20 nodes. 
For generating solutions, instead of generating programs independently at random (which results in a vast majority of duplicates), we generate the programs that are most likely using the search algorithm of Menon et al.~\citeyear{menon2013machine}, which does not produce duplicates. Since problem generation is not a bottleneck, problems are generated more simply by sampling uniformly from the grammar $\mathbb{P}$ and then removing the duplicates. 

\subsubsection{Domains.} We consider CFGs from the following domains: 
\begin{itemize}[leftmargin=*,noitemsep,topsep=0pt]
\item \emph{Number Theory.} Example target problems include GCD or finding the largest non-trivial factor of a (small) number, using a brute force approach. Numerous trivial functions also arise in the practice set, such as given two numbers, output the smaller of the two.
\item \emph{Finding Roots.} Example target problems are finding the root of algebra expressions such as $\log(y/2) - x^2=0$, i.e., solve for $y$ as a function of $x$.
\item \emph{Summation Formulas.} Example target problems are finding the closed-form expression of computing the sum of a function of the first $n$ numbers $f(n)=\sum_{i=1}^n g(i)$. For example $\sum_{i=1}^n i^2 = n(n+1)(2n+1)/6$.
\item \emph{Strings}. For simplicity, we considered problems where the desired output is a single character from the input meeting a certain objective. An example problem is finding the most frequent character in a string or finding the alphabetically first character in a string. 
\end{itemize}
\begin{figure*}[ht]
\centering
\subfloat{\includegraphics[scale=0.18]{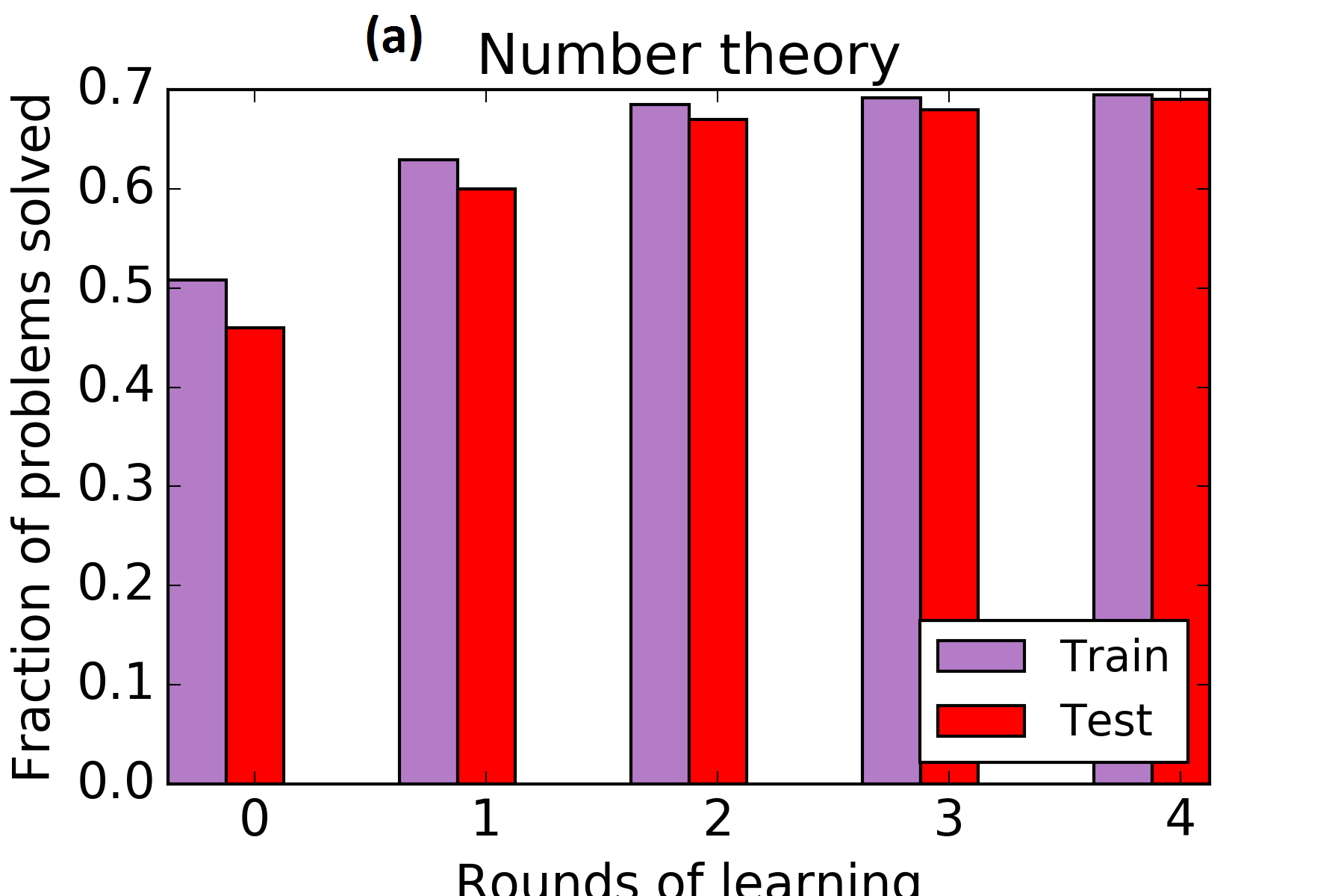}}
\subfloat{\includegraphics[scale=0.18]{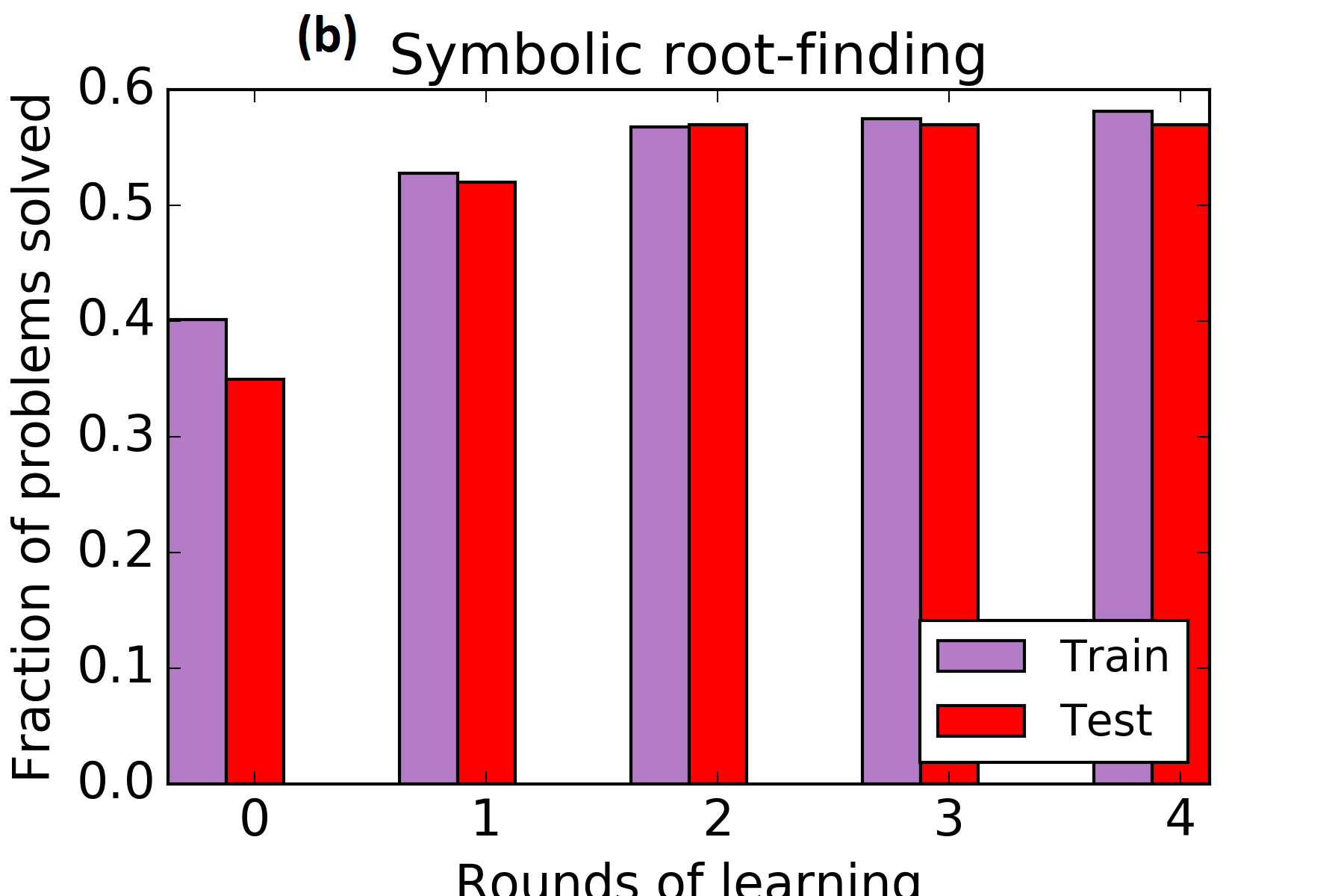}}
\subfloat{\includegraphics[scale=0.18]{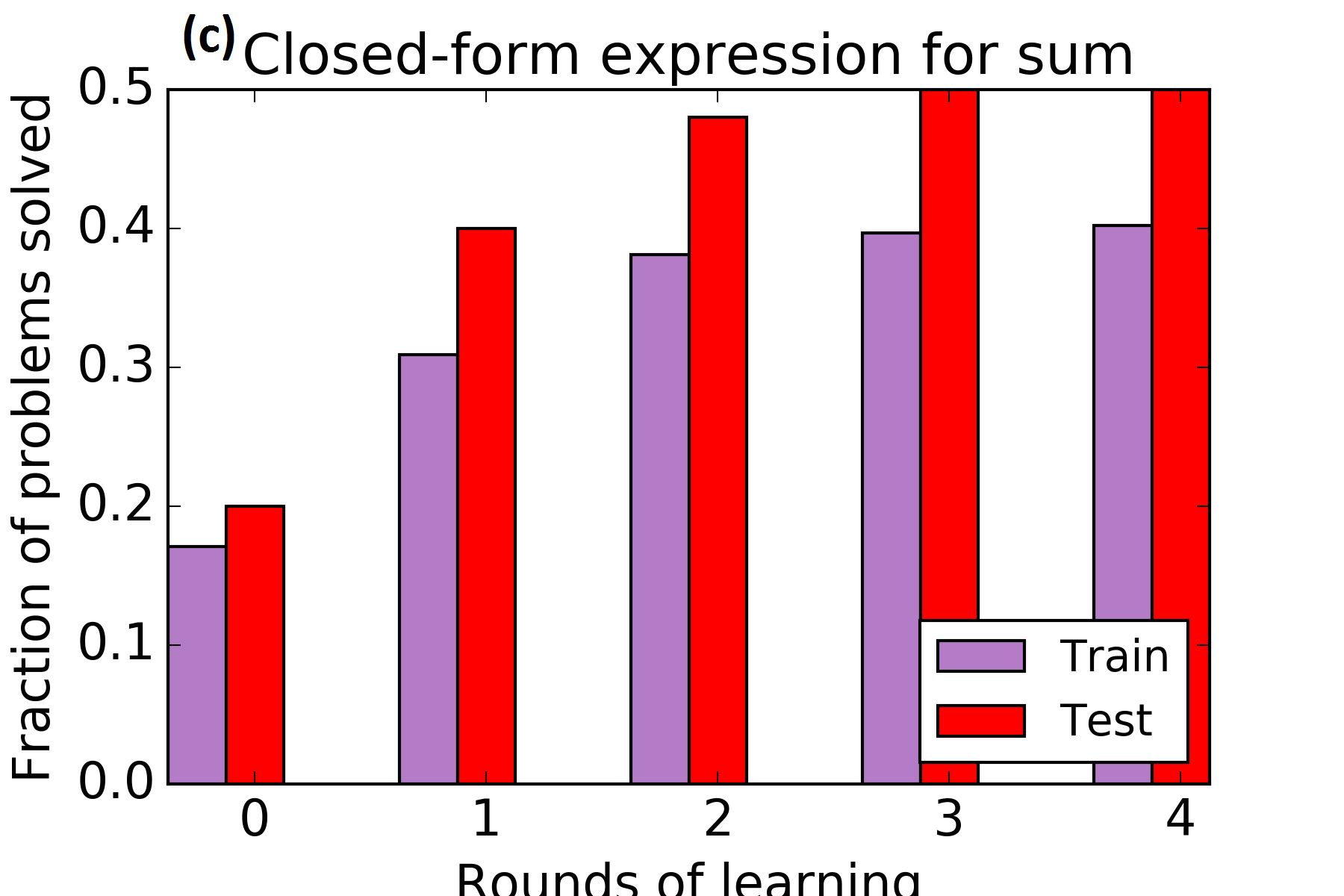}}
\subfloat{\includegraphics[scale=0.18]{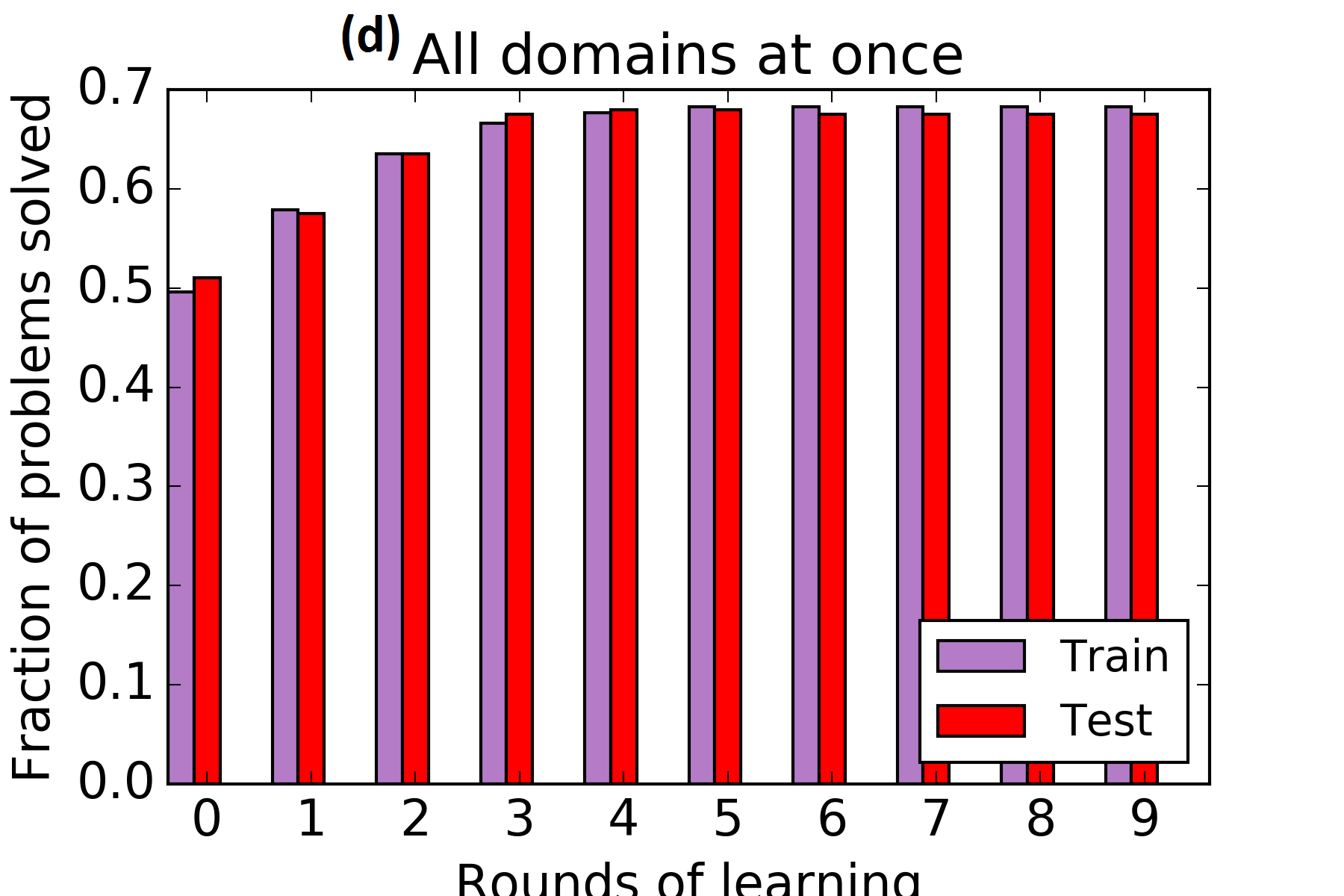}}
\caption{(a-c) Domain-specific experiment: Fraction of problems solved with multiple rounds of learning in each domain (every training round was run for 45 minutes). 
 (d) All-domain experiment: Similar results are achieved with 90 minutes (instead of 45) for each learning round, and additional rounds beyond 5 do not seem to improve test accuracy.}
\label{fig:exp1}
\end{figure*}

\begin{table*}[th]
\centering
\begin{tabular}{|l|lllll|}
\hline
Challenge problem number &  1 & 2 & 3 & 4 &  5  \\
Solution time without learning & $>$30:00 & $>$2:15 & $>$2:15 & 0:05 &  2:15 \\
Solution time with all-domain learning & $>$2:15 & $>$2:15 & 0:01 & 0:01 &  0:01\\
Solution time with domain-specific learning & 0:10 & $>$2:15 & 0:01 & 0:01 & 0:01\\
\hline
\hline
Challenge problem number &   6 & 7 &8 &9 & 10  \\
Solution time without learning &  $>$2:15& $>$2:15 &$>$2:15&$>$2:15 & $>$2:15 \\
Solution time with all-domain learning & 0:03 & $>$2:15 & 0:01 & $>$2:15 & 0:01\\
Solution time with domain-specific learning & 0:01 & 0:05 & 0:01 &$>$2:15 & 0:01 \\
\hline
\end{tabular}
\caption{Time in hours and minutes for solving each of the 10 challenge problems with and without learning. Most problems were run for a max of 2:15, but GCD was run for 30 hours without learning.}
\label{table:continuous}
\end{table*}

\noindent \textbf{Verification.} For each of these problems, the performance is evaluated on a bounded domain $X$ of possible inputs and range $Y$ of possible outputs, which makes it easy to check for optimality, and enables early stopping for efficiency. 

\noindent \textbf{Grammars.} Briefly, our grammars, both for our typed program-scoring grammar $\mathbb{P}$ and of our solution grammar $\mathbb{S}$, include operators such as $+$, $-$, $*$, $/$, $mod$, math functions such as \texttt{pow}, \texttt{abs}, \texttt{tan}, \texttt{tanh}, \texttt{arctanh}, \texttt{log}, \texttt{exp}, constants such as 0, 1, 2, or the passed arguments, operators such as $<$, \texttt{max}, boolean operators such as \texttt{or}, \texttt{if} expressions,  string functions such as \texttt{startswith}, \texttt{endwith}, \texttt{count}, \texttt{upper}, \texttt{lower}, \texttt{ord}. Additionally, the grammar $\mathbb{S}$ contains recursive functions (with a single, or two arguments passed), operators on lists, sets and tuples.

Each generated program was converted to a string and passed to the native python eval. A timeout of 1 second was used per evaluation. 



\subsection{Results on Practice/ Test Problems}

To empirically evaluate our proposed framework, we conducted experiments under two experimental conditions. In the first condition, we considered for the  context-free grammars, only the one relevant to the problems we were trying to solve. For example, for solving string problems, we considered the string CFG, and so on. In the second
condition we took the union of all grammar rules from our four different domains, and considered 250 problems from each of the four domains. We refer to this as the \emph{all-domain experiment}. 

For the first condition, we considered 1,000 practice train problems and 100 test problems. We progressively performed four training rounds, as more rounds did not seem to improve learning.  Each round of training was run for 45 minutes. We show in Figure \ref{fig:exp1}(a-c) the fraction of  practice train/test problems solved by GlassPS as rounds of learning progress from 0 to 4. We can see that performing 4 rounds of our framework  typically solves about 70\% of the target problems. The string domain is not shown because 100\% of the problems were solved in each round, even prior to learning. 

For the second setup of the all-domain experiment, we show the results in Figure \ref{fig:exp1}(d). In order to achieve similar fraction of train/test problems solved as in the first setup, every training round was run for double time, i.e., 90 mins instead of 45. We conclude that although better results can be achieved with domain-specific learning, the all-domain experiment is a proof of concept experiment that even without knowing the domain, GlassPS can improve itself over time. 
 
\subsection{Results on Challenge problems}
\label{subsec:challenges}
Finally, we consider the performance of our framework in our following ten challenge problems. 

\begin{enumerate}[noitemsep]
\item 
GCD (number theory), 
\item 
Greatest non-trivial factor of a number $n$.\footnote{The time limits were such that a brute-force loop could find the largest factor in the alloted time, so no advanced factoring algorithms were necessary.} (number theory),
\item 
Most frequent character in a string (strings), 
\item 
Alphabetically first character in a string (strings),
\item 
Solve for $y$: $\log(y) - (x * x) / 2=0$ (roots), 
\item 
Solve for $y$: $y + (\text{pow}(x, (2 * 2)) / 2)=0$ (roots),
\item 
Find a closed-form expression for: sum1(1, n, lambda i: (i * i)) (sums),  
\item 
Find a closed-form expression for: sum1(1, n, lambda i: (i *(i * i))) (sums),  
\item 
Find a closed-form expression for: sum1(1, n, lambda i: pow(2, (-i))) (sums),
\item 
Find a closed-form expression for: sum1(1, n, lambda i: 1 / (1, (i * (1 + i)))) (sums). 
\end{enumerate}

In Table \ref{table:continuous}, we report the time in hours and minutes needed to solve each of the 10 challenge problems with our system. We can compare the times needed for (i) when the search is not guided by learning, i.e., brute-force (2nd row), (ii) the all-domain setup (3rd row), and (iii) domain-specific learning (4th row), which is  the best performing one across problems. 

Notably, our framework can solve problems that would be otherwise prohibitively slow without learning. Notably, GCD was found after 30 hours with no learning, while it took only 10 minutes using domain-specific learning, and 2 hours and 15 minutes for the all-domain experiment.


%% file: sec-concl.tex
In this paper, we introduced a general framework for learning to write computer programs that maximize the score of glass-box objective functions. 
We have formulated this as a machine learning problem, showing as a proof-of-concept that a simple classifier such as logistic regression, can successfully learn patterns among the features of the scoring programs, and the features of the generated solution programs to be scored.
We have shown experimentally that our framework learns over time to generate python typed code that solves problems across domains, even though we did not provide types and we did not give access to the domain type each problem comes from. 
While the learning could certainly be improved (e.g. using deep learning), the approach is shown to be sound and to some extent flexible enough to combine multiple domains of synthesis in a single system. This opens up interesting directions for multi-domain learning systems that learn to solve problems. \\\\

\noindent \textbf{Acknowledgements.} The authors would like to thank Huseyn Melih Elibol for the useful discussions.
